\documentclass[10pt,conference]{IEEEtran}
\IEEEoverridecommandlockouts                            
 
\usepackage{url}
\usepackage[bookmarks=false]{hyperref}
\usepackage{graphicx}
\usepackage{enumerate}
\usepackage{amsfonts}
\usepackage{amsmath}
\usepackage{algorithm2e}
\usepackage{algorithmic}
\usepackage{listings}
\usepackage[table]{xcolor}
\usepackage{soul}
\usepackage{multirow}
\usepackage[numbers]{natbib}
\usepackage{tikz}
\usetikzlibrary{shapes}
\usetikzlibrary{arrows}

\definecolor{codegreen}{rgb}{0,0.6,0}
\definecolor{codegray}{rgb}{0.5,0.5,0.5}
\definecolor{codepurple}{rgb}{0.58,0,0.82}
\definecolor{backcolour}{rgb}{0.95,0.95,0.92}

\lstdefinestyle{mystyle}{
    backgroundcolor=\color{backcolour},   
    commentstyle=\color{codegreen},
    keywordstyle=\color{magenta},
    numberstyle=\tiny\color{codegray},
    stringstyle=\color{codepurple},
    basicstyle=\ttfamily\footnotesize,
    breakatwhitespace=false,         
    breaklines=true,                 
    captionpos=b,                    
    keepspaces=true,                 
    numbers=left,                    
    numbersep=5pt,                  
    showspaces=false,                
    showstringspaces=false,
    showtabs=false,                  
    tabsize=2
}

\title{On Improving Deep Learning Trace Analysis with System Call Arguments}

\author{\IEEEauthorblockN{Quentin Fournier}
\IEEEauthorblockA{\textit{Polytechnique Montréal}\\
Quebec H3T 1J4 \\
quentin.fournier@polymtl.ca}
\and
\IEEEauthorblockN{Daniel Aloise}
\IEEEauthorblockA{\textit{Polytechnique Montréal}\\
Quebec H3T 1J4 \\
daniel.aloise@polymtl.ca}
\and
\IEEEauthorblockN{Seyed Vahid Azhari}
\IEEEauthorblockA{\textit{Ciena}\\
Ottawa K2K 0L1 \\
vazhari@ciena.com}
\and
\IEEEauthorblockN{François Tetreault}
\IEEEauthorblockA{\textit{Ciena}\\
Ottawa K2K 0L1 \\
ftetreau@ciena.com}
}

\makeatletter \def\@IEEEpubidpullup{8\baselineskip} \makeatother

\begin{document}

\maketitle
\thispagestyle{empty}
\pagestyle{empty}

\begin{abstract}
Kernel traces are sequences of low-level events comprising a name and multiple arguments, including a timestamp, a process id, and a return value, depending on the event. Their analysis helps uncover intrusions, identify bugs, and find latency causes. However, their effectiveness is hindered by omitting the event \emph{arguments}. To remedy this limitation, we introduce a general approach to learning a representation of the event names along with their arguments using both embedding and encoding. The proposed method is readily applicable to most neural networks and is task-agnostic. The benefit is quantified by conducting an ablation study on three groups of arguments: call-related, process-related, and time-related. Experiments were conducted on a novel web request dataset and validated on a second dataset collected on pre-production servers by Ciena, our partnering company. By leveraging additional information, we were able to increase the performance of two widely-used neural networks, an LSTM and a Transformer, by up to 11.3\% on two unsupervised language modelling tasks. Such tasks may be used to detect anomalies, pre-train neural networks to improve their performance, and extract a contextual representation of the events.\let\thefootnote\relax\footnotetext{\copyright 2021 IEEE. Personal use of this material is permitted. Permission from IEEE must be obtained for all other uses, in any current or future media, including reprinting/republishing this material for advertising or promotional purposes, creating new collective works, for resale or redistribution to servers or lists, or reuse of any copyrighted component of this work in other works.\\To appear in: Proceedings of the 18th International Conference on Mining Software Repositories (MSR ’21), Madrid, Spain.}
\end{abstract} 


\section{Introduction}
\label{sec:introduction}

In recent years, deep learning has been successfully applied to an ever-growing range of supervised and unsupervised tasks. This trend has been enabled by the ever-increasing computational resources and the novel techniques introduced to take advantage of these resources. As of today, the largest model for natural language processing (NLP) comprises 175 billion parameters and has been trained on half a terabyte of curated text \citep{2020arXiv200514165B}. The authors showed that the model performance scales consistently with the number of parameters and the amount of available data.

\begin{figure*}[!t]
    \centering
    \includegraphics[width=\linewidth]{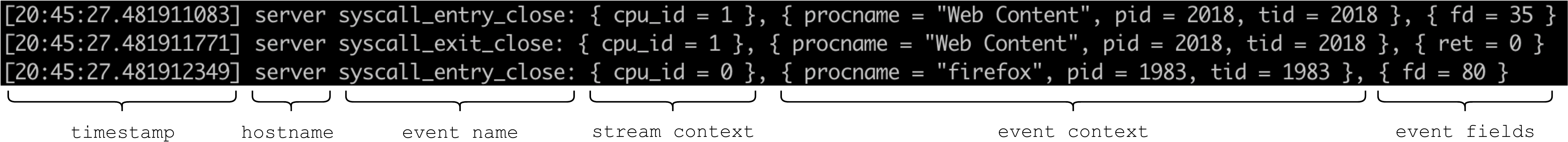}
    \caption{Trace events as displayed by Babeltrace\protect\footnotemark. The arguments are all the values except the event name. In this example, the arguments are from left to right: the timestamp, the hostname, the CPU id, the process name, the process id, the thread id, the file descriptor, and the return value.}
    \label{fig:events}
\end{figure*}

A technique that surely generates a large amount of data is \textit{tracing}. Tracing is the act of collecting a trace which is a sequence of low-level events. Such events are produced whenever a specific instruction called tracepoint is encountered at runtime and comprises a name, a precise timestamp, and possibly many arguments. Figure~\ref{fig:events} depicts three trace events.
 
Traces provide insights on the execution of a piece of code and have been extensively used to detect intrusions, identify bugs, and find the root cause of latency issues. The main advantages of tracers are (1) they do not require to stop the execution contrary to debuggers, and (2) they do not aggregate events or metrics contrary to loggers.

In this paper, we consider the events generated by the operating system also known as kernel events. The benefit of such events is two-fold: (1) tracepoints are already implemented in the Linux kernel, which allows tracing virtually any Linux system without having to modify the source code, and (2) the behaviour of the whole system is visible from the kernel. In this paper, we focus on a subset of the kernel events called \textit{system calls}. System calls are the only way for an application to communicate with the operating system.

Although the manual inspection of traces may reveal insights that are virtually impossible to extract automatically, the amount of human labour required is often prohibitive. Indeed, the operating system produces thousands of events every second, most of which may be collected. The sheer size of traces is the primary reason why automatic analysis is required. Traces are used to detect \textit{unknown} intrusions, to identify \textit{unknown} bugs, or to locate the \textit{unknown} root cause of anomalies, making their analysis often challenging to specify in practice. Therefore, machine learning techniques, that is, techniques that learn how to solve a task from examples, are well suited to analyse traces.

Most machine learning methods take a vector of numerical features as input. Hand-crafted features of traces have been proposed, but no representation seems to work universally well or to encapsulate the true underlying explanatory factors \citep{Murtaza2013, 8790013, fournier19}. Instead of relying on hand-crafted features, neural networks learn how to extract meaningful features for the task. By finding a relevant input representation for the task, neural networks reduce the need for an expert, and the model performance is improved in most cases.

Although a wide range of deep learning techniques has been applied on traces by previous works, only a small fraction of the accessible information has been considered. The event arguments and, in certain cases, the event ordering inside the trace, have been left out in the literature. Section~\ref{sec:litterature} discusses in more detail the related works and their limitations. We argue that the increase in resources and the improvement of deep learning techniques allow fully exploiting traces.

A trace is a sequence of discrete values arguably comprising a syntax and a semantic. Due to their resemblance to natural language, the most common approach is to apply deep learning techniques from natural language processing. Our methodology follows the previous works by considering the widely used Long Short-Term Memory (LSTM)~\cite{lstm}. A recent alternative to LSTM for processing variable-length sequences called the Transformer \cite{NIPS2017_7181} is also evaluated. Although this model is omnipresent in NLP, it has not yet been applied on traces. The two models were evaluated on two unsupervised objectives: (1) left-2-right language model (LM) that allows computing the likelihood of a sequence, and therefore, detecting anomalies, and (2) masked language model (MLM) that is used for pre-training \cite{Devlin2018}.

This paper's first contribution is the introduction in Section~\ref{sec:method} of a novel method to learn a single representation of the system call names with their arguments. Results are detailed in Section~\ref{sec:results} and an ablation study is conducted to investigate the impact of three groups of arguments: call-related, process-related, and time-related.

The second contribution of this paper is the introduction of a novel dataset comprising around 250,000 web requests. The actual dataset is provided, but most importantly, the data generation methodology is explained in Section~\ref{sec:data}. One may argue that our dataset is too simple or that it inaccurately represents actual web servers. Therefore, every experiment is validated on a second dataset collected on pre-production servers by Ciena, the partnering company of this research.

Finally, Section~\ref{sec:threats} discusses the possible threats to validity, and Section~\ref{sec:conclusion} answers interesting questions about the pertinence of the proposed approach and future works.

\footnotetext{\url{https://babeltrace.org}}

\section{Related Work}
\label{sec:litterature}

Over the last two decades, a wide range of machine learning techniques has been applied to analyze traces, including naive Bayes \citep{Asmitha2014}, random forest \citep{song:hal-02264598}, and hidden Markov models \cite{6566581}. Recently, the trend has shifted toward more flexible approaches, and especially toward deep learning methods. Model flexibility relates to the space of functions that the model is able to learn and increases with the number of parameters. Therefore, highly flexible methods, such as large neural networks, are able to learn complex solutions that typically perform better than less flexible ones. This section provides an overview of the main neural networks that have been studied in the tracing literature as well as their limitations.

\begin{figure}[!htb]
	\centering
	\includegraphics[width=0.85\linewidth]{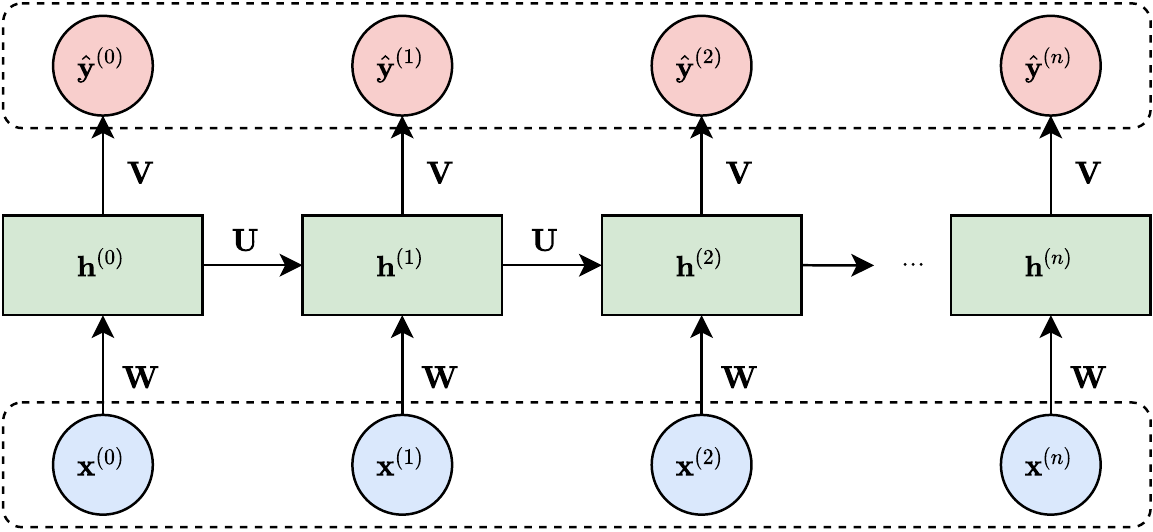}
	\caption{The unrolled computational graph of a recurrent neural network. The input and output sequences are depicted in blue and red, respectively. The time step is indicated in exponent and between parenthesis. Note that the network parameters $W$, $U$, and $V$, are replicated at every time step. Therefore, the network can process variable-length input sequences.}
	\label{fig:rnn}
\end{figure}

Recurrent neural networks (RNNs) allow processing variable-length sequences with a fixed number of parameters. Such a network produces an output at every time step and is depicted in figure \ref{fig:rnn}. The Long Short-Term Memory (LSTM) \citep{lstm} is a recurrent neural network specifically designed to learn dependencies across a large number of time steps. This network has been extensively and successfully used across many fields. Tracing is no exception, and LSTM is by far the most popular neural network to analyze traces \citep{ song:hal-02264598, DBLP:journals/corr/DymshitsMT17, Kim2016, 8814585, 9025172}.

\citet{DBLP:journals/corr/DymshitsMT17} trained a unidirectional and a bidirectional LSTM on \textit{sequences of system call count vectors}. Such vectors are bag-of-words, that is to say, the normalized counts of system call names, from a fixed-duration window. This aggregation is a trade-off between computational efficiency and performance, and is controlled by the window size. The authors also trained an Inception-like net consisting of multiple LSTMs with tied weights. They found that simpler LSTMs performs on par with the more complex ones.

\citet{Kim2016} trained an ensemble of LSTMs on sequences of system call names. Ensemble techniques improve the performance, although not significantly, and the robustness of the chosen method. While ensemble techniques may be necessary for industry products, this paper will not leverage them as the main objective is to show the relative impact of the arguments rather than the approach's absolute performance.

\citet{song:hal-02264598} compared an LSTM with less flexible machine learning techniques to detect and explain anomalies from streams of traces. They did not, however, explicitly say which events were considered or describe their preprocessing.

\begin{figure}[!htb]
	\centering
	\includegraphics[width=\linewidth]{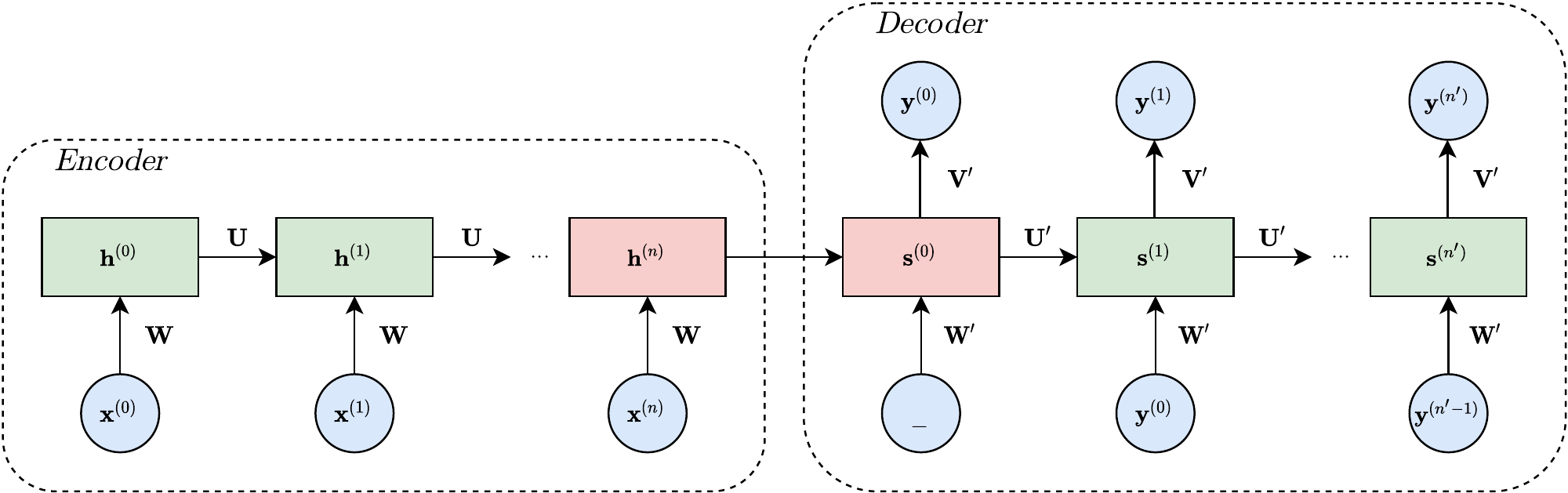}
	\caption{Sequence-to-sequence framework. A first network (encoder) encodes the input sequence into a fixed-size context $\textbf{h}^{(n)}$ shown in red, then a second network (decoder) generates the output sequences based on this context.}
	\label{fig:seq-to-seq}
\end{figure}

Recurrent neural networks output a vector at every time step, so the output sequence must have the same length as the input sequence (see Figure \ref{fig:rnn}). This property of RNNs may become a constraint depending on the task. To overcome this limitation, \citet{NIPS2014_5346} introduced the sequence-to-sequence framework where a first network (encoder) encodes the input sequence into a fixed-size context. A second network (decoder) then generates the output sequences based on this context. This framework allows outputting a variable-length sequence independently of the input sequence length and is illustrated in Figure~\ref{fig:seq-to-seq}.

\citet{DBLP:journals/corr/abs-1808-01717} used a gated recurrent unit\footnote{GRU is similar to LSTM but requires fewer parameters.} (GRU) \cite{DBLP:journals/corr/ChoMGBSB14} in a sequence-to-sequence fashion to extend sequences of system calls names and increase the accuracy of intrusion detection. 

\begin{figure}[!htb]
	\centering
	\includegraphics[width=0.7\linewidth]{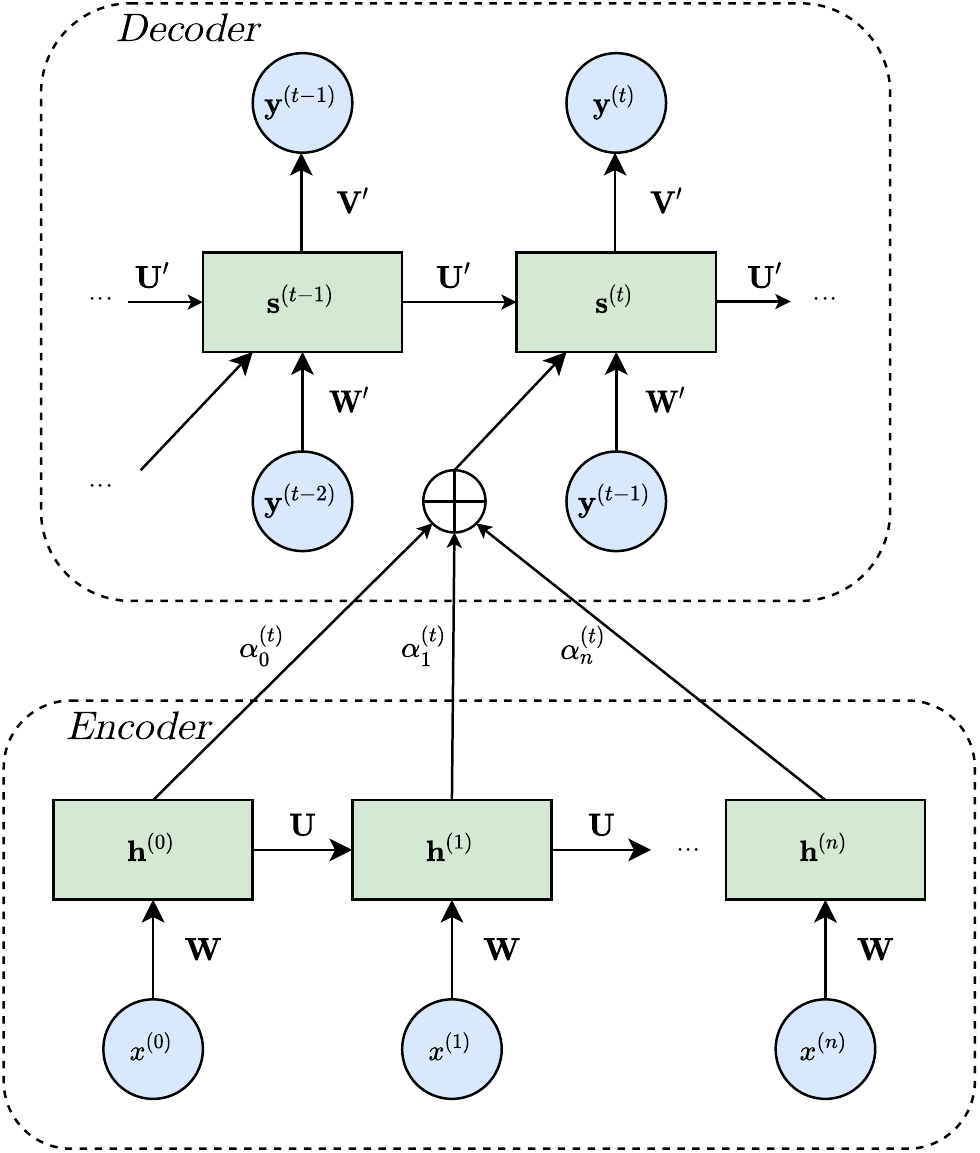}
	\caption{Inter-attention mechanism. The attention weight $\alpha_i^{(t)}$ corresponds to the strength with which the $i$-th encoder hidden representation $h^{(i)}$ contributes to the context of the $t$-th decoder step.}
	\label{fig:attention}
\end{figure}

Recurrent networks, including LSTMs and GRUs, suffer from an issue related to memory compression \cite{DBLP:journals/corr/ChengDL16}. As the input sequence gets processed, information must be stored in the fixed-size hidden representation $\boldsymbol{h}$. Either $\boldsymbol{h}$ is too large and computational resources are wasted, or $\boldsymbol{h}$ is too small and information is lost. In the latter case, the model performance might be significantly impacted. \citet{Bahdanau2014} introduced an alignment mechanism called \textit{inter-attention} to mitigate the effect of memory compression. This mechanism computes a different representation of the input for each output step, effectively allowing the decoder to ``look at'' the relevant part(s) of the input for each output step. Figure \ref{fig:attention} illustrates the inter-attention mechanism.

\citet{DBLP:journals/corr/abs-1803-04967} augmented an LSTM with the dot-product inter-attention and explored different ways of computing the attention: fixed, position-based (syntax attention), and context-based (semantic attention). For the task of system log anomaly detection, every attention yielded comparable results.

Finally, due to their sequential nature, recurrent networks do not scale efficiently to longer sequences \citep{NIPS2017_7181}. \citet{DBLP:journals/corr/abs-1901-02860} introduced the relative effective context length (RECL), the largest context length that leads to a substantial relative gain over the best model. Simply put, increasing the context length over the RECL yields a negligible increase in performance; thus, RECL indicates the maximum dependency length that the model is able to learn. They showed that the RECL of LSTM is limited to around 400 time-steps. This is problematic for trace analysis since hundreds of events may be generated every second.

To overcome this limitation, \citet{NIPS2017_7181} introduced the Transformer, a sequence-to-sequence model based solely on the inter-attention and self-attention mechanisms. The self-attention allows relating any two positions in a sequence regardless of their distance thus allowing for a significant increase in performance in most natural language processing tasks at the cost of a quadratic complexity with respect to the sequence length. To the best of our knowledge, this model has not been applied on traces but has been included considering its ubiquity in NLP.

None of the aforementioned works considered the system call arguments. Arguably, the main reason is that the community does ``\textit{[...] not have a compact fixed-dimensional representation for system call arguments suitable for large-volume training and classification.}'' \citet{DBLP:journals/corr/DymshitsMT17}.

Nonetheless, \citet{8814585} used a bimodal LSTM that is the concatenation of two LSTM hidden representations trained on the real-valued duration and one-hot-encoded texts, respectively. Albeit their work considered logs rather than traces, one may view their method as leveraging a temporal argument. The neural network proposed by \citet{9025172} is the closest to actually considering multiple arguments values. The authors trained an LSTM using the system call name, the CPU cycles count, and the distribution of characters in the arguments' values.

As far as we know, only two works by \citet{doi:10.1142/S0218213006003028, DBLP:conf/flairs/TandonC05} considered the actual values of multiple system call arguments. The authors trained a conditional rule-learning algorithm called LERAD, which, contrary to neural networks, does not require to learn a fixed-size representation of the arguments.

\section{Proposed Approach}
\label{sec:method}

Before introducing the proposed approach, let us clarify the different categories of system call arguments. One may group them depending on whether they are part of the stream context, the event context, or the event fields (see Figure~\ref{fig:events}). In this work, the arguments are grouped based on their semantic. The first category comprises all \textit{call-related} arguments such as the return value, the file descriptor, the type of futex operation, and the number of bytes to write -- depending on the event. The second category consists of all \textit{process-related} arguments such as the process name, the thread id, and the process id. Note that this category corresponds exactly to the event context. Finally, the third group consists of \textit{time-related} arguments such as the timestamp and the timeout duration.

The scope of this work is limited to the arguments that are common to virtually all system calls. Namely, the return value (\texttt{ret}), whether the event corresponds to the start or end of a system call execution (\texttt{entry}), the process name (\texttt{procname}), the thread id (\texttt{tid}), the process id (\texttt{pid}), and the timestamp (\texttt{timestamp}). As explained later, extending this work to other arguments is simple but may require a substantially larger dataset. Table~\ref{tab:args} recapitulates the considered arguments.
\setlength\tabcolsep{5pt}
\def\arraystretch{1.25}
\begin{table}[!htb]
\centering
\caption{The studied system call arguments.}
\resizebox{\linewidth}{!}{\begin{tabular}{lllll}
Category             & Argument         & Notation    & Type \\ \hline
\multirow{2}{*}{call-related} & return value & \texttt{ret}    & integer \\
                 & start/end of execution & \texttt{entry}  & boolean \\ \hline
\multirow{3}{*}{process-related} & process name       & \texttt{procname} & string \\
                 & process id        & \texttt{pid}   & integer \\
                 & thread id         & \texttt{tid}   & integer \\ \hline
time-related         & timestamp  & \texttt{timestamp} & integer \\ \hline
\end{tabular}}
\label{tab:args}
\end{table}

In order to determine how to represent the arguments, one must identify the intrinsically meaningful ones. In other words, one has to assess whether the argument values convey meaning in themselves -- without any context. As an example, let us consider the process name ``apache''. This value means that an Apache web server has generated the system call, hence \texttt{procname} is inherently meaningful. On the contrary, the process id ``12523'' is only meaningful in the context of the trace. Indeed, the \texttt{pid} allows relating events that have been generated by the same process; the value ``12523'', however, may well be associated with two distinct processes at different points in time.

The \texttt{procname}, the \texttt{return value}, and the \texttt{entry} are intrinsically meaningful arguments, and hence, an embedding will be learned for them. On the contrary, the \texttt{pid}, the \texttt{tid}, and the \texttt{timestamp} are not inherently meaningful and an encoding will be applied.

\subsection{Embedding}

One way to represent textual words is through a sparse binary vector called one-hot-encoding. The $i$-th word of the vocabulary is mapped to a row vector $\boldsymbol{e}_{w_i}$ whose dimension is equal to the size of the vocabulary. Such vector is filled with $0$ except for the $i$-th position which is equal to $1$. Given a toy vocabulary of three system call names $\{\texttt{open}, \texttt{close}, \texttt{timer}\}$, their one-hot-encoding would be $[1, 0, 0]$, $[0, 1, 0]$, and $[0, 0, 1]$, respectively.

One-hot-encoding has two major drawbacks: (1) the vector dimension is equal to the vocabulary size which may be large, and (2) the encoding of any two distinct words are perpendicular, meaning that words are equidistant. For instance, one would expect $\text{dist}(\boldsymbol{e}_\texttt{open},\boldsymbol{e}_\texttt{close})<\text{dist}(\boldsymbol{e}_\texttt{open},\boldsymbol{e}_\texttt{timer})$ as \texttt{open} is semantically closer to \texttt{close} than to \texttt{timer}.

A better representation is expected to be more compact and to encapsulate semantic knowledge about the word. Such representation is called an \textit{embedding}. Note that in the natural language processing community, an embedding refers to both the general mapping from a textual space to a semantic vector space and the actual dense vectorial representation of a word.

Formally, an embedding is defined by a dense matrix $\boldsymbol{W} \in \mathbb{R}^{d_v \times d_e}$ with $d_v$ the size of the vocabulary and $d_e$ the dimension of the embedding such as $d_e \ll d_v$. The embedding $\boldsymbol{x}_{w_i}$ of the word $w_{i}$ is computed by multiplying its one-hot-encoding $\boldsymbol{e}_{w_i}$ with the embedding matrix $\boldsymbol{W}$ which effectively acts as a lookup table (see example below). The embedding matrix is typically treated as any other model parameter in that it is randomly initialized and learned with gradient descent.

\begin{equation*}
\resizebox{.95\hsize}{!}{$
    \underbrace{[\begin{array}{cccc}
    0 & 0 & \cellcolor{yellow} 1 & 0
    \end{array}]}_{\text{One-hot vector }\boldsymbol{e}_{w_i}}
    \times
    \underbrace{\left[\begin{array}{ccccc}
    5 & 6 & 2 & 1 & 4\\
    0 & 1 & 7 & 3 & 1\\
    \rowcolor{yellow} 4 & 8 & 1 & 6 & 9 \\
    3 & 1 & 2 & 8 & 2 \\
    \end{array}\right]}_{\text{Embedding matrix }\boldsymbol{W}}
    =
    \underbrace{[\begin{array}{ccccc}
    4 & 8 & 1 & 6 & 9
    \end{array}]}_{\text{Word embedding } \boldsymbol{x}_{w_i}}$}
\end{equation*}

\subsection{Encoding}

It would be ill-advised to learn an embedding of a value that is not inherently meaningful -- whose interpretation depends entirely on the context. Instead, one should use a deterministic transformation without any parameter that is called an encoding.

Once more, let us consider the process id. Neural networks take as input a vector of numerical values. Therefore one may provide the actual \texttt{pid} as input. It is, however, a best practice to normalize the input vector to mitigate numerical instabilities, help training, and improve the model performance. Since the \texttt{pid} is not inherently meaningful in general\footnote{There are exceptions. Notably, \texttt{pid} 0, \texttt{pid} 1, and kernel-reserved \texttt{pid}s are meaningful and could be considered separately.}, any bijection from the argument space to a small interval such as $[0, 1]$ or $[-1, 1]$ works well. The simplest solution would be to map the \texttt{pid} uniformly to real values between $[0, 1]$. In practice, the number of distinct \texttt{pid} within a trace varies and is often unknown beforehand.

A practical way to encode a numerical value is to apply the cosine function. Indeed, the codomain is $[-1,1]$, and the function requires no knowledge about the distribution or the extremum of the input variable. The cosine function is not, however, a bijection. As a result, collisions may occur: two different values assigned to the same encoding. Consider $x=1$ and $x'=1+4\pi$:
\begin{equation*}
  cos(1)=cos(1+4\pi)
\end{equation*}

The number of collisions may be reduced by dividing $x$ by an appropriately large number which effectively controls the period of the cosine function. Note that if the denominator is too small, collisions may still occur.
\begin{equation*}
  cos(1/2)=cos((1+ 4 \pi)/2)
\end{equation*}

If the denominator is too large, the encodings will be extremely close, hence difficult for a model to distinguish.
\begin{align*}
  cos(1/1000)\approx 0.9999995\\[0.5em]
  cos((1+ 4 \pi)/1000)\approx 0.99991
\end{align*}

Instead, the denominator should be equal to the estimated maximum value that $x$ can take. 

The number of collisions may be further reduced by applying multiple cosine functions with different periods. In that case, the encoding is a vector comprising the output of each cosine function. In other words, the output of every cosine function is concatenated into a vector which is the encoding.

Our approach relies on the encoding proposed by~\citet{NIPS2017_7181} which leverages an alternation of cosine and sine functions with an increasing denominator. More formally, the encoding of a numerical value $x$ is a vector $\boldsymbol{pe}_x$ of dimension $d$ whose $j$-th value is given either by equation~\ref{eq:pe1} or~\ref{eq:pe2} depending on whether $j$ is even ($j=2i$) or odd ($j=2i+1$), respectively.
\begin{align}
 pe_{x,2i} &= sin(x/10000^{2i/d})\label{eq:pe1}\\[0.5em]
 pe_{x,2i+1} &= cos(x/10000^{2i/d})\label{eq:pe2}
\end{align}

\begin{figure}[!htb]
  \centering
  \includegraphics[width=0.8\linewidth]{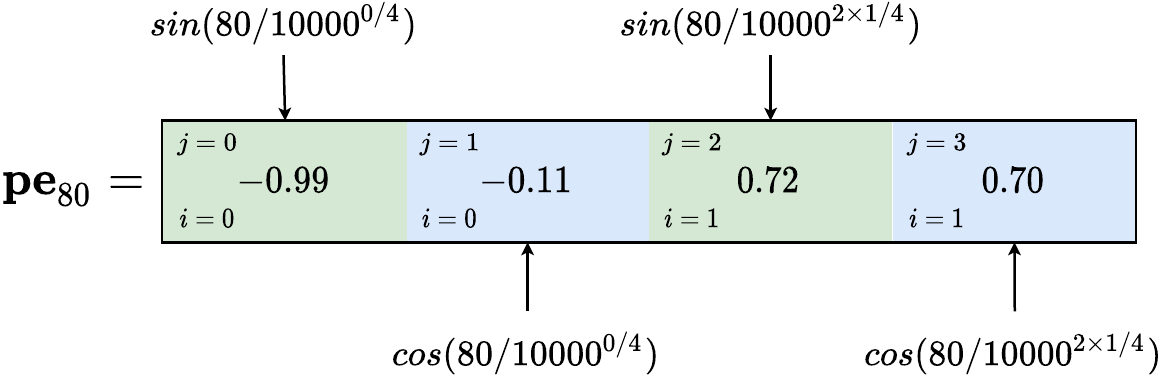}
  \caption{Encoding of the value $x=80$ using~\citet{NIPS2017_7181} formula and a dimension $d=4$.}
  \label{fig:encoding}
\end{figure}

As the authors underlined, there exists a linear relation between the $\boldsymbol{pe}_x$ and $\boldsymbol{pe}_{x+k}$, which they hypothesized should facilitate learning. Figure~\ref{fig:encoding} illustrates the encoding.

\subsection{Addition or Concatenation}

Let us now investigate how to combine the arguments embedding and encoding into a single event representation. The two most common approaches are the addition and the concatenation.

\begin{figure}[!ht]
	\centering
	\includegraphics[width=0.85\linewidth]{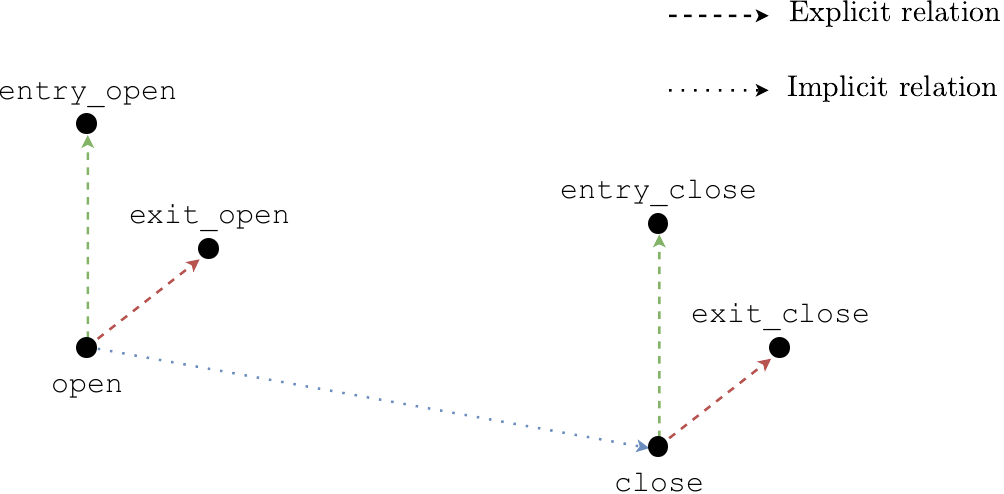}
	\caption{The embedding of the system call names ``open'' and ``close''. The green and red dashed lines represent the explicitly modelled ``entry'' and ``exit'' relations, respectively. The blue dotted line represents the implicitly modelled relation between the two system call names.}
	\label{fig:addition}
\end{figure}

One may describe the addition of two vectors $\boldsymbol{x}$ and $\boldsymbol{y}$ as the translation of a point $\boldsymbol{x}$ by a vector $\boldsymbol{y}$ -- or equally a point $\boldsymbol{y}$ by a vector $\boldsymbol{x}$. Let us consider the system call name and the argument \texttt{entry} which has two possible values, ``entry'' and ``exit''. Furthermore, let us consider the system call name embedding as a point and the \texttt{entry} embedding as a vector. The addition effectively shifts the system call name embedding depending on whether the event corresponds to the start or the end of the system call execution. As illustrated by figure~\ref{fig:addition}, the relation has been explicitly modelled in the same space as the embedding of the system call name, which simplifies their visualization and interpretation.

The addition preserves the dimension, which may be too small to store all the information, thus creating a bottleneck. Instead, the concatenation allows combining vectors without such a bottleneck. Indeed, the dimension of the resulting vector is the sum of the dimensions of the concatenated vectors. That may, however, be a drawback if the model size scales with the input dimension as larger models are computationally expensive to train and prone to overfitting. One may mitigate the overfitting by collecting a sufficiently large dataset.

Although the embedding visualization is outside of the scope of this work, we believe interesting to model the system call name, the argument \texttt{entry}, and the argument \texttt{ret} in the same space. One may gain insights into the system by investigating the relations between those vectors. Therefore, only those values will be added, and the remaining arguments will be concatenated. Note that it would be ill-advised to add an encoding to an embedding since the former is not inherently meaningful.

\subsection{Event Representation}

\begin{figure*}[!t]
	\centering
	\includegraphics[width=0.75\linewidth]{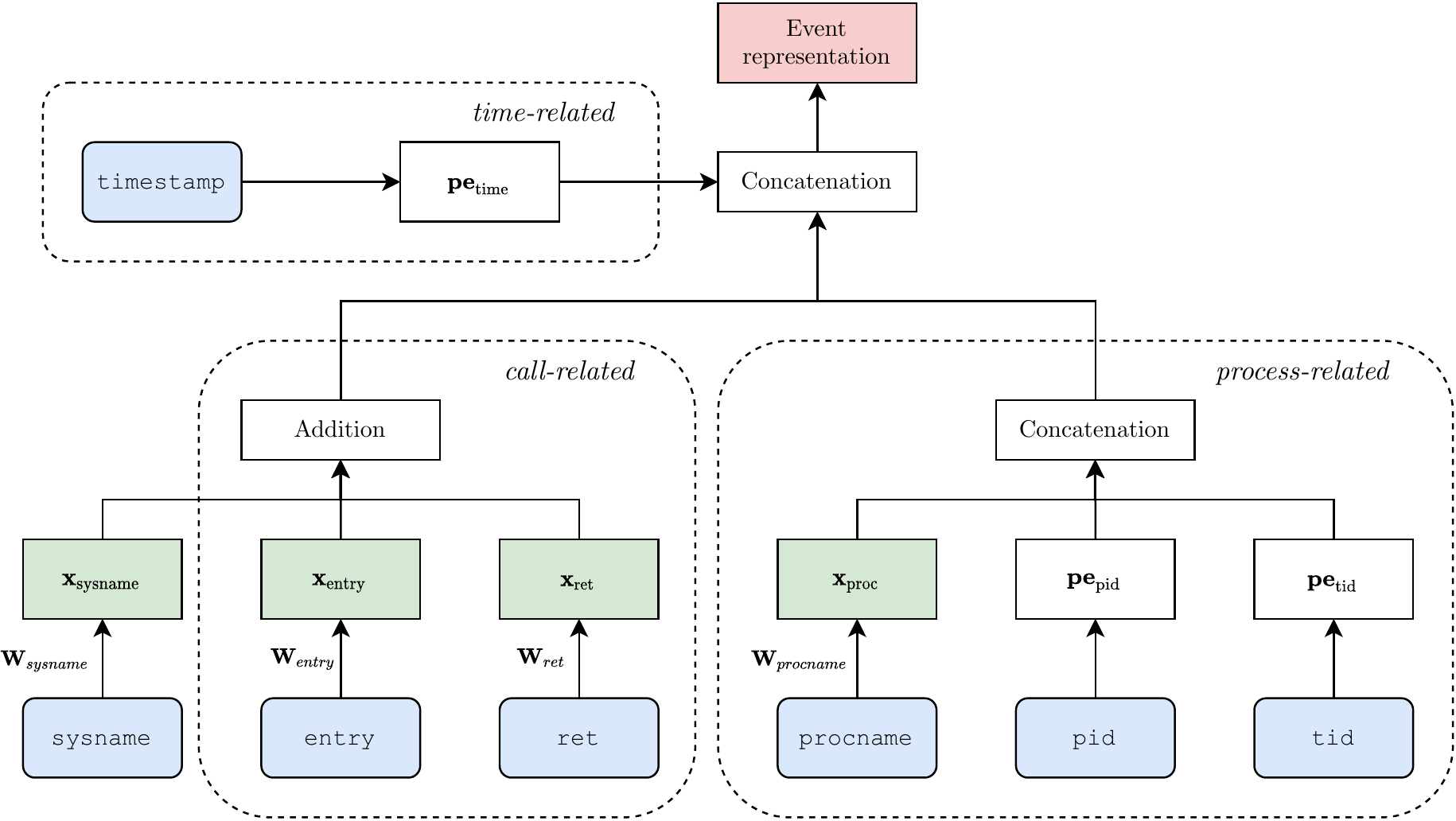}
	\caption{Computational graph of the event representation. Blue rounded rectangles represent the arguments. Green rectangles indicate that the transformation is learned (embedding), and the parametrization is noted next to the incoming arrow. White rectangle indicates that the transformation is not learned (encoding, addition, or concatenation).}
	\label{fig:embedding}
\end{figure*}

Figure~\ref{fig:embedding} illustrates the computational graph of the event representation. For the call-related arguments, the embedding of the \texttt{sysname}, the \texttt{entry}, and the \texttt{ret} are added. Note that the return value is simplified to either ``success'' if the numerical value is greater or equal to zero, or ``failure'' otherwise. For the process-related arguments, the \texttt{procname} embedding is concatenated with the \texttt{pid} and \texttt{tid} encodings.
For the time-related argument, the \texttt{timestamp} is converted from nanoseconds to microseconds and is encoded. Finally, the representation of each category of arguments is concatenated. 

Neural networks take numerical values as input that may be arranged as vectors, matrices, or, more generally, tensors. In the case of traces, the network's input is typically a sequence of vectors corresponding to the events. Such vectors may be the one-hot encoding of system call names, or better, their embedding. The proposed approach outputs a vectorial representation of the event with its arguments; therefore, it applies to most deep learning models.

The proposed event representation is non-contextual: a system call with its arguments will have the same representation regardless of the other trace events. Some tasks greatly benefit from a contextual representation which may be obtained with a Transformer trained on the masked language model objective~\citep{Devlin2018}. Although such a model has been evaluated, contextual representations are outside the scope of this paper.

\section{Data Collection}
\label{sec:data}

Over the years, many tracing datasets have been explored; however, most of them are not publicly available. Consequently, the now-obsolete UNM~\citep{unm} and KDD98~\citep{kdd98} datasets are still widely used~\citep{Murtaza2013}. Those datasets were collected more than two decades ago and are clearly not representative of modern systems anymore. Therefore, they should not be used to evaluate recent approaches. In 2013, \citet{Murtaza2013} and \citet{conf/wcnc/CreechH13} addressed this issue by introducing two new datasets: FirefoxDS and ADFA-LD, respectively. Unfortunately, the former is unavailable, and the system call arguments were omitted from the latter.

\begin{figure}[!htb]
	\centering
	\includegraphics[width=\linewidth]{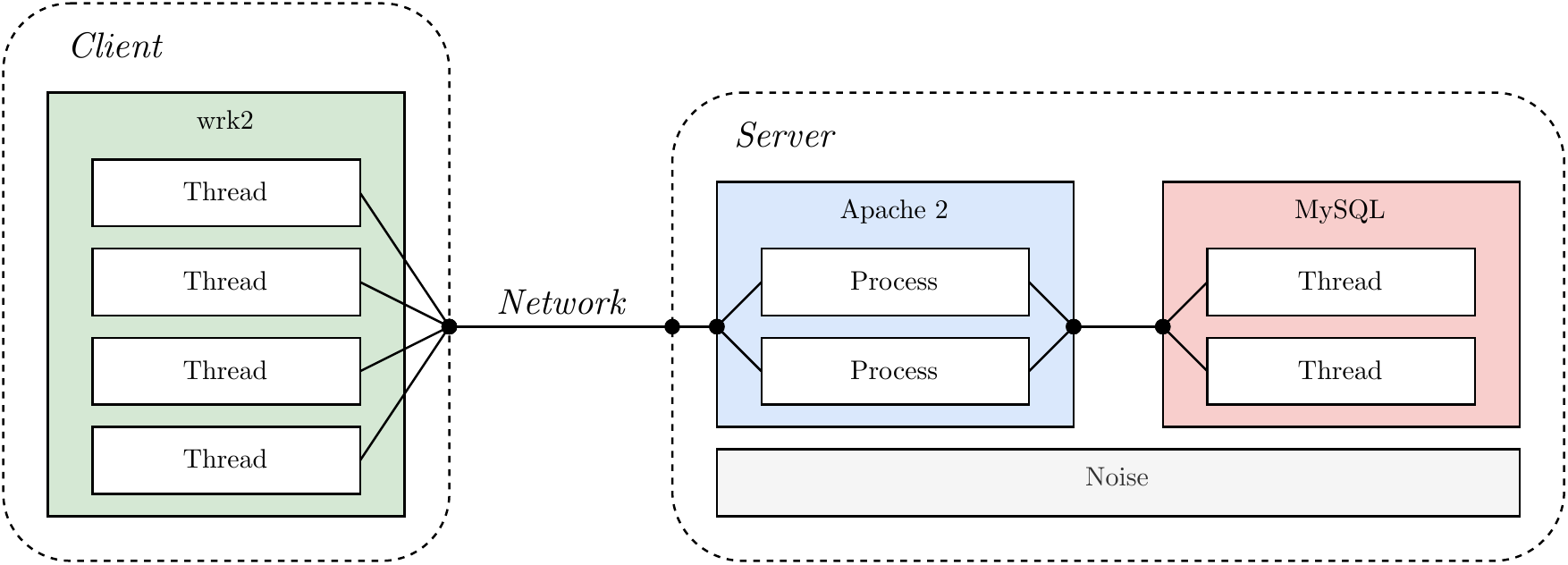}
	\caption{Client-server framework. The client and the server are two distinct physical machines that communicate over the network. The server may be executing other software while handling a request which is considered to be noise from a request point of view.}
	\label{fig:client-server}
\end{figure}

As indicated by~\citet{2020arXiv200514165B}, increasing the size of language models greatly improves their performance regardless of the task. As larger models require more data to be properly trained, the dataset must not only be modern but also massive. To the best of our knowledge, no massive and modern datasets comprising the system call arguments are publicly available. To that extent, we propose to generate such a dataset using \emph{requests}. A request is a task delimited by specific start and end events. Examples include database queries, micro-services, and application functions. Notably, web requests have been extensively studied in the literature as they are ubiquitous. We introduce a methodology to generate a massive dataset of web request traces using a simple client-server framework (see Figure~\ref{fig:client-server}). The source code and the dataset are publicly available on GitHub\footnote{
\url{https://github.com/qfournier/syscall_args}
} and Zenodo\footnote{
\url{https://zenodo.org/record/4091287\#.X4hhGNjpNQI}
}, respectively.

\subsection{Methodology}

On the client-side, a benchmark tool is used to send many concurrent requests to the server via the hypertext transfer protocol (HTTP). We chose \texttt{wrk2}\footnote{\url{https://github.com/giltene/wrk2}}, an open-source multithreaded equivalent of the Apache benchmark, as it guarantees a constant throughput load with an accuracy up to 99.9999\% for sufficiently long runs. Moreover, \texttt{wrk2} yields a latency summary which allows extracting statistics about the dataset without processing it.

On the server-side, a web server handles the client requests and communicates with a database to retrieve the necessary information. For the web server, we chose Apache2 for its omnipresence and its modularity. Indeed, Apache2 is the most popular web server since 1996, and its vast community has developed many optional modules, including app servers and database connection managers. For the database, we chose MySQL for its ease of use and performance. MySQL is filled with the Sakila Sample Database\footnote{\url{https://dev.mysql.com/doc/sakila}} which includes an \texttt{author} table comprising ids, first names, and last names. Finally, PHP was installed as an Apache module to query the database.

One may be interested in simulating different behaviours such as slow or abnormal requests. In order to increase the likelihood of such requests, the server must be overloaded, which is done by restricting the amount or speed of the resources (CPU, memory, network, and disk). Consequently, Apache2 is deployed in a virtual machine using Virtual Box.

Physical servers often execute multiple tasks simultaneously. Since our server was dedicated, Firefox was automatically and randomly called from the console to take screenshots of random Wikipedia pages. The monitoring tools \texttt{htop} and \texttt{bmon} were also running in separate terminals. This allows creating a load on the CPU, the disk, and the network, as well as generating random events in the trace which adds variability.

In this work, we focus on the server-side since it is the source of most delays. A single trace is collected during the entire benchmark, therefore containing many individual requests. Depending on the task at hand, one may consider the whole trace as a single sequence or individual requests as separate sequences. Several tracers are available; however, the \textit{Linux Tracing Toolkit: next generation} (LTTng) \citep{lttng} is often the prefered choice given its lightweight and rapidity. Although only some system calls arguments are considered in this work, all arguments have been collected in order to have a complete view of the system.

\subsection{Dataset Analysis}

The server was deployed on a virtual machine with two cores from an Intel Core i7-8700 (up to 4.6 GHz), 1 Gb of DDR4 RAM, and an NVME SSD. The operating system was Ubuntu 18.04. Different throughputs were used to simulate different usages: idle, low, medium, and high. High usage means that the server is barely able to handle requests in real-time and that some end up with a timeout. Note that the training set and the test set were collected separately using different throughputs to avoid any overlap.

We collected around 250,000 requests which amount to almost 150 million system calls. One would likely have to collect a larger dataset in order to consider additional arguments such as the file descriptor without overfitting.

\begin{figure}[!htb]
  \centering
  \includegraphics[width=\linewidth]{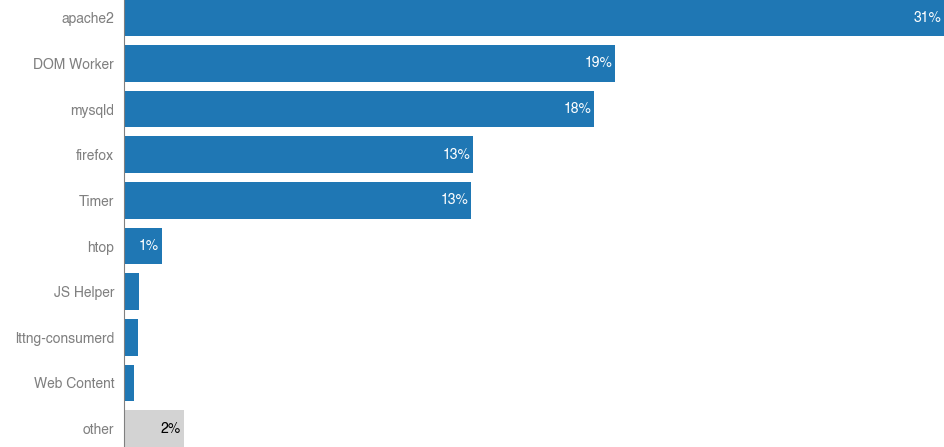}
  \caption{Distribution of process names}
  \label{fig:dist_proc}
\end{figure}

Figure \ref{fig:dist_proc} depicts the distribution of process names. As expected, the three most frequent processes are those that handle requests, namely the web server, its workers, and the database. Note that Firefox is responsible for issuing 13\% of the system calls.

\begin{figure}[!htb]
  \centering
  \includegraphics[width=\linewidth]{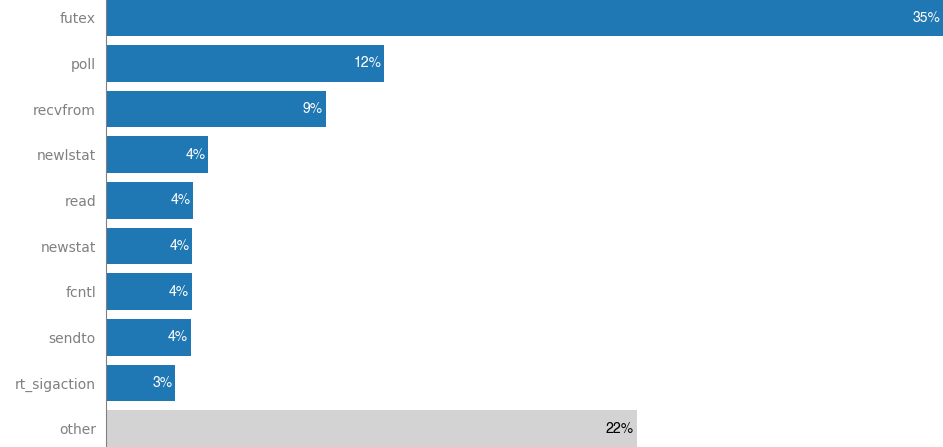}
  \caption{Distribution of system calls}
  \label{fig:dist_syscall}
\end{figure}

Figure \ref{fig:dist_syscall} depicts the distribution of system call names. The two most frequent system calls are \texttt{futex} and \texttt{poll} which provide a method for waiting until a condition becomes true and until a file descriptor becomes available to perform IO operations, respectively. This behaviour is to be expected in networked multicore systems, especially when many remote requests are being handled concurrently.

For an equivalent analysis of Ciena's dataset, we refer the reader to the GitHub repository.

\section{Computational Experiments}
\label{sec:results}

This section introduces the neural networks and objectives on which the system call arguments' impact was evaluated. The source code, hyperparameters, and trained models are publicly available on GitHub\footnote{\url{https://github.com/qfournier/syscall_args}}.

\subsection{Networks}

The first model evaluated is a deep unidirectional Long Short-Term Memory (LSTM) network with two hidden layers comprising 96 units. The vast majority of existing works to analyze traces apply an LSTM on system call names only~\citep{song:hal-02264598, DBLP:journals/corr/DymshitsMT17, Kim2016, 8814585, 9025172}; therefore, those methods would require almost no modification to leverage the arguments with the proposed approach.

The second model evaluated is a Transformer. Transformers are highly parallelizable and are able to learn dependencies across an unlimited number of steps at the price of quadratic complexity. Many works address this limitation; however, since this paper aims to demonstrate the usefulness of the system call arguments, we settled for the vanilla Transformer introduced by \citet{NIPS2017_7181}. In particular, the network consists of six layers, each comprising 8 attention heads and a feedforward network with 128 units. 

Contrary to LSTMs, Transformers are agnostic to the event position in the sequence. To solve this shortcoming, \citet{NIPS2017_7181} injected positional knowledge by summing a positional encoding with the embedding. In our experiments, the model achieved better results when the positional encoding was concatenated to the event embedding.

The dimensions of the arguments embedding and encoding have a significant impact on the model performance; thus, various configurations were evaluated. The following dimensions performed well in all experiments: 32 for the \texttt{sysname}, \texttt{entry}, and \texttt{ret}, 16 for the \texttt{procname}, 4 for the \texttt{pid} and \texttt{tid}, and 8 for the \texttt{timestamp}. Consequently, the dimension of the whole event representation is 64. Note that the dimension of the positional encoding was equal to that of the \texttt{timestamp}.

\subsection{Objectives}

The first objective is the left-to-right language model (LM), which predicts the conditional probability of the next system call name given the previous system calls. The chain rule allows computing the joint probability of the whole sequence, that is, its likelihood, and therefore may be used to detect changes in the system behaviour, intrusions, and anomalies. Notably, \citet{Kim2016} used language modelling for host-based intrusion detection.

\begin{figure}[!htb]
    \centering
    \includegraphics[width=\linewidth]{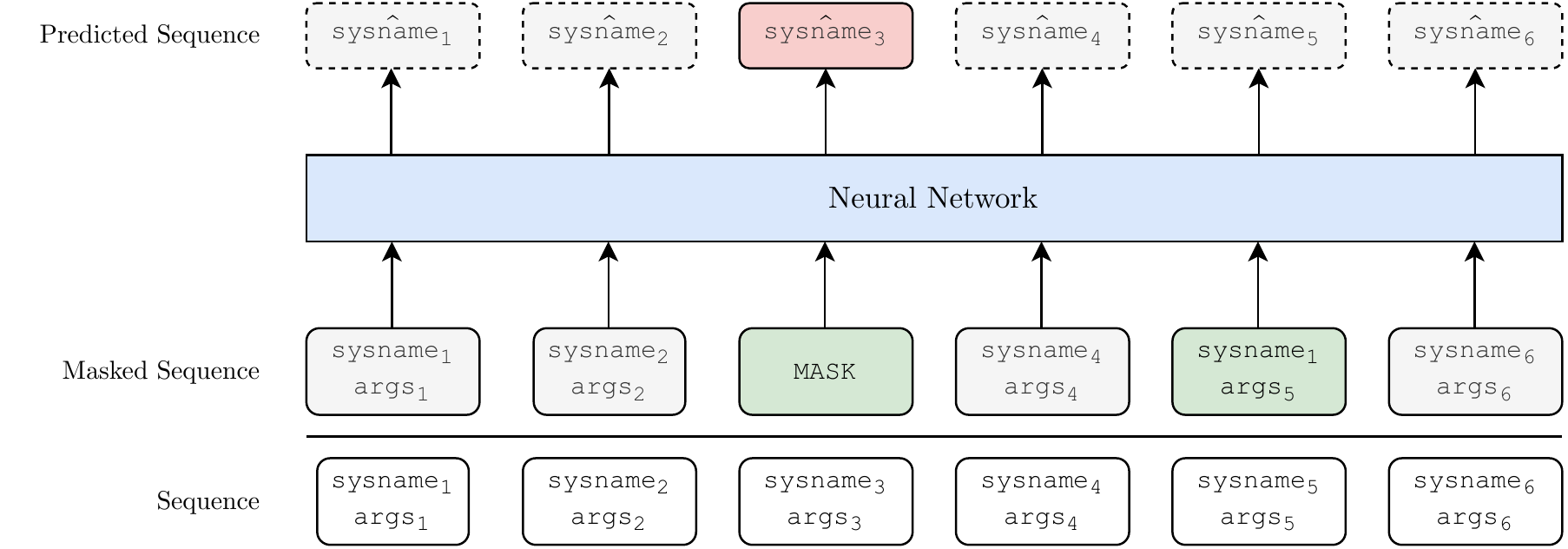}
    \caption{Masked language model. Events in green have been randomly selected, and system call names in red are the predictions independently considered.}
    \label{fig:mlm}
\end{figure}

The second objective is the masked language model (MLM), which independently estimates the probability of masked words given the rest of the sequence. The more events are masked, the less context is available, and the more difficult is the training. In practice, MLM is often used to pre-train neural networks, and it has been shown to improve the model performance on downstream tasks, that is, the tasks of interest. Therefore, we evaluated the pre-trained model on LM in a zero-shot manner and determined that masking 25\% of the events performed reasonably well on both datasets (see Table \ref{tab:mlm}). In particular, we followed the methodology of \citet{Devlin2018} by randomly selecting 25\% of the events, of which 80\% were entirely masked, 10\% were replaced by a random system call name with the same argument values, and 10\% were left unchanged. Randomly replacing the selected events generates noise which increases the robustness of the model. The proportion of random events is identical to \citet{Devlin2018} as their ablation study showed it worked well for pre-training. Note that masked LMs are technically not language models as they are not trained to maximize the joint probability of sentences. Figure~\ref{fig:mlm} illustrates the masked language model. 

\subsection{Data}

Due to memory constraints on the graphics processing unit (GPU), the models must be trained on small sequences. Therefore, the traces were split into non-overlapping sequences of 256 events. Note that those sequences do not correspond to requests. One would need to implement the proposed approach with a lower computational complexity model in order to process whole requests as they usually contain thousands of events.

The first dataset studied has been introduced in Section~\ref{sec:data} and comprises 318,674 training sequences and 258,190 test sequences. The second dataset has been collected by Ciena on pre-production servers executing proprietary software and comprises 190,924 training sequences and 64,628 test sequences. Although smaller, this dataset is designed to be representative of a real-world use case.

A quarter of each test set was randomly selected to create a validation set on which the hyperparameters were fine-tuned, and the model was evaluated at train time for early stopping.

\subsection{Results}

For each combination of datasets, objectives, and neural networks, two event representations have been compared: the system call name without any argument (\texttt{none}) and with every argument as described in Figure~\ref{fig:embedding} (\texttt{all}). 

Arguments may affect the performance differently; however, the computational cost of evaluating the impact of each argument, or worse, each combination of arguments, is prohibitive in practice. Instead, we evaluated the global impact of three groups of arguments: call-related (\texttt{entry} and \texttt{ret}), process-related (\texttt{procname}, \texttt{pid}, and \texttt{tid}), and time-related (\texttt{timestamp}).

Because the arguments embedding and encoding are concatenated, considering additional arguments increases the event representation dimension, which also increases the model size. On one side, the additional information allows the network to be larger without overfitting; hence one may see the increase in size as a byproduct of the arguments. On the other side, one may argue that a larger model only considering the system call name would perform better. In order to test these hypotheses, a compensated model considering no argument is evaluated (\texttt{none cmp.}). The dimension of the \texttt{sysname} embedding is increased from 32 to 64, which is the event representation's dimension when all the arguments are considered.

The model performance was measured in terms of cross-entropy (lower is better) and top-1 accuracy (higher is better). The cross-entropy is a measure of the difference between two distributions, in our case, the model output and the label. In the usual case of one-hot labels, the cross-entropy is defined as the negative logarithm of the correct event's predicted probability. The top-1 accuracy is the percentage of correct predictions, where a prediction is the system call name with the highest predicted probability. Results are detailed in Table~\ref{tab:results}.

\setlength\tabcolsep{10pt}
\def\arraystretch{1.70}
\begin{table*}[!htb]
\caption{Impact of three categories of system call arguments (cross-entropy/accuracy).}
\label{tab:results}
\centering
\begin{tabular}{ccccccccc}
                                                             &                     &             & \texttt{none}         & \texttt{none cmp.}    & \texttt{time}         & \texttt{call}         & \texttt{process}      & \texttt{all}          \\ \hline
\multirow{3}{*}{\shortstack{Web \\ Requests}} & \multirow{2}{*}{LM} & LSTM        & 0.528 / 83.1 & 0.529 / 83.1     & 0.526 / 83.2 & 0.451 / 85.6 & 0.443 / 85.7   & \textbf{0.423} / \textbf{86.4} \\
                              &                     & Transformer & 0.609 / 80.3 & 0.506 / 83.3     & 0.599 / 80.6 & 0.489 / 84.3 & 0.452 / 85.0   & \textbf{0.380} / \textbf{87.3} \\ \cline{2-9} 
                              & MLM                 & Transformer & 0.535 / 81.7 & 0.485 / 82.8     & 0.524 / 81.8 & 0.400 / 87.2 & 0.423 / 85.0   & \textbf{0.182} / \textbf{94.1} \\ \hline
\multirow{3}{*}{Ciena}        & \multirow{2}{*}{LM} & LSTM        & 0.294 / 91.8 & 0.301 / 91.5     & 0.301 / 91.6 & 0.277 / 92.2 & 0.283 / 91.9   & \textbf{0.264} / \textbf{92.4} \\
                              &                     & Transformer & 0.323 / 90.4 & 0.292 / 91.3     & 0.310 / 90.8 & 0.290 / 91.5 & 0.271 / 91.9   & \textbf{0.238} / \textbf{92.8} \\ \cline{2-9} 
                              & MLM                 & Transformer & 0.285 / 90.8 & 0.264 / 91.3     & 0.270 / 91.2 & 0.202 / 94.0 & 0.245 / 91.8   & \textbf{0.125} / \textbf{96.2} \\ \hline
\end{tabular}
\end{table*}

In every experiment, the models that consider all the arguments achieved the lowest cross-entropy and the highest accuracy. The compensated models perform on par or better than their smaller counterpart; however, they are systematically outperformed by the models considering all the arguments. These results indicate that the increase in performance is not only due to the increase in model size but also to the additional arguments. Therefore, the arguments must contain useful information for language modelling tasks. Interestingly, the masked language model objective benefits more from call-related arguments than process-related ones.

The time-related argument has a negligible impact on the LSTMs; consequently, the temporality must be of little use for the left-to-right language model objective. Nonetheless, Transformers appear to benefit from the \texttt{timestamp} and, as a result, an ablation study of the \texttt{timestamp} and the \texttt{position} was conducted to quantify their impact. The results shown in table~\ref{tab:order} reveal that \texttt{timestamp} does increase the performance over a model without any arguments, although not as much as the \texttt{position}, which indicates that Transformers are able to leverage the redundancy of the positional information embedded in the \texttt{timestamp}. However, with an equal number of parameters, a model considering only the \texttt{position} performs on par or better than one considering both values. Such behaviour is to be expected since the positional information in the timestamp is harder to extract. One may be tempted to dismiss the \texttt{timestamp}; however, it should be noted that some downstream tasks, including latency detection, may greatly benefit from the \texttt{timestamp}.

\begin{table}[!htb]
\centering
\caption{Impact of the event's position and timestamp encoding dimensions on the Transformer without arguments (cross-entropy/accuracy). A dimension of zero is equivalent to omitting the argument.}
\begin{tabular}{cccc}
\texttt{timestamp} & \texttt{position} & Web Requests & Ciena \\ \hline
0 & 0 & 0.730 / 76.9 & 0.444 / 86.9 \\
8 & 0 & 0.661 / 78.4 & 0.337 / 89.8 \\
0 & 8 & 0.609 / 80.3 & 0.323 / 90.4 \\
8 & 8 & 0.599 / 80.6 & \textbf{0.310} / \textbf{90.8} \\
0 & 16 & \textbf{0.587} / \textbf{80.9} & 0.313 / \textbf{90.8} \\ \hline
\end{tabular}
\label{tab:order}
\end{table}

As shown in Table \ref{tab:overhead}, the computational overhead imposed by the additional arguments was negligible compared to the overall training cost, making the proposed approach suitable for real-world applications. This is to be expected as the embedding is simply a matrix multiplication, and the encoding is only a small number of cosine and sine functions.

\begin{table}[!htb]
    \centering
    \caption{Average epochs time ($\pm$ std) in milliseconds of the Transformers trained on the web requests.}
    \begin{tabular}{lcc}
         & LM & MLM\\\hline
        \texttt{none}       & 99.3 ($\pm$ 2.0)  & 232.2 ($\pm$ 8.2)\\
        \texttt{none cmp.}  & 102.2 ($\pm$ 1.6) & 232.8 ($\pm$ 5.3)\\
        \texttt{time}       & 104.1 ($\pm$ 2.6) & 228.5 ($\pm$ 3.8)\\
        \texttt{call}       & 102.4 ($\pm$ 2.1) & 227.0 ($\pm$ 6.6)\\
        \texttt{process}    & 103.6 ($\pm$ 1.7) & 234.3 ($\pm$ 6.9)\\
        \texttt{all}        & 106.0 ($\pm$ 2.4) & 238.5 ($\pm$ 4.5)\\\hline
    \end{tabular}
    \label{tab:overhead}
\end{table}

\begin{table}[!htb]
    \centering
    \caption{Impact of the percentage of selected events for pre-training the Transformer with all arguments as evaluated on LM (cross-entropy/accuracy).}
    \begin{tabular}{ccc}
    $p_{mask}$ & Web Requests & Ciena       \\ \hline
    0.05     & 3.826 / 54.6 & 1.738 / 80.1  \\
    0.10     & 3.881 / 55.7 & 1.641 / 80.2  \\
    0.15     & \textbf{3.314} / \textbf{56.7} & 1.639 / 80.2  \\
    0.20     & 3.543 / 56.1 & 1.617 / \textbf{80.4}  \\
    0.25     & 3.334 / 56.1 & 1.647 / \textbf{80.4}  \\
    0.30     & 3.387 / 56.3 & \textbf{1.548} / 80.2  \\ \hline
    \end{tabular}
    \label{tab:mlm}
\end{table}

\section{Threats to Validity}
\label{sec:threats}

The main threat to validity is the limited scope of the evaluation. Indeed, the approach has only been evaluated on two unsupervised language modelling tasks due to the lack of a publicly available dataset comprising the system call arguments. To mitigate this limitation, we provide the source code as well as the trained models for researchers and practitioners to evaluate our approach to their task.

The second threat to validity is the simplicity of the environment on which our dataset was collected. Consequently, the dataset may not represent real-world use cases and may not reflect the approach's actual benefit. This limitation is addressed by evaluating the two objectives on a second dataset collected by Ciena on pre-production servers. Additionally, our dataset is unlabelled. Consequently, it is challenging to use for supervised tasks such as anomaly detection. To alleviate this shortcoming, we provide a tutorial and the scripts required to generate the dataset such that users can produce their own labels.

% Another limitation of our dataset\footnote{Daniel: Here, it is not clear whether you are talking about our dataset or/and our dataset  collected by Ciena. If you are talking about the first, you might want to talk about this limitation (i.e., lack of labels) before} 

Finally, although the proposed approach's computational overhead is negligible, neural networks still require powerful GPUs to be trained. The models' average training time described in Table~\ref{tab:results} was less than 2 hours, with the slowest model taking about 5 hours on a single NVIDIA RTX2080Ti and two Intel Xeon Bronze 3104 1.7Ghz. Therefore, the experiments are easily reproducible with modest computational resources.

\section{Concluding Remarks}
\label{sec:conclusion}

In practice, it is often difficult to determine whether a specific deep learning approach is beneficial for the task at hand. In this section, we answer two general questions to help researchers and practitioners decide whether to adopt the proposed method.

\textit{Do the arguments invariably increase the model performance?} We argue that the performance either improves or remains the same, provided two conditions. Firstly, the model must be flexible enough to be able to extract relevant information from the arguments. Such a model would be able to leverage the additional information in order to make more informed predictions, hence more accurate. If the arguments only contain irrelevant information to the task, the performance cannot increase. It may, however, decrease. Indeed, larger inputs translate into larger embeddings, which increase the model size, hence its flexibility. As the model flexibility increases, it becomes prone to overfitting, that is, to learn peculiarities from the dataset that do not reflect real explanatory factors. It is well-known that the difference between training and generalization errors grows with the model flexibility and shrinks with the number of training examples~\citep{Goodfellow-et-al-2016}. Therefore, the second condition is that enough samples must be available to prevent the model from overfitting. Large datasets of traces are typically easy to obtain, so the amount of data is not a limiting factor. Notably, this work introduced a methodology to generate a massive dataset of requests. Furthermore, many techniques such as dropout~\citep{srivastava2014dropout}, batch normalization~\citep{ioffe2015batch}, and early stopping~\citep{Caruana2000OverfittingIN} allow mitigating the overfitting that may occur. Nonetheless, the arguments should be omitted if one knows beforehand that the information is irrelevant to the task. For instance, if a single thread is recorded, the \texttt{tid} is constant and may be safely omitted.

\textit{In practice, how does one know when to consider additional arguments?} It seems that one would need to estimate a priori (1) if the model is complex enough, (2) if the dataset is large enough, and (3) if the arguments could be relevant to the task at hand. Fortunately, in the case of neural networks, the models are generally more flexible than necessary -- they contain many more parameters than there are samples in the dataset~\citep{Caruana2000OverfittingIN}. As explained above, collecting large datasets of traces is often trivial, and the risk of overfitting may be significantly reduced. When possible, we recommend considering the arguments and comparing the model with a baseline that does not.

In this work, we introduced a massive dataset of web requests and a general approach to learning a representation of the system call names along with their arguments. By leveraging the left-out information, we were able to systematically increase the performance of two neural networks on two language-modelling tasks at a negligible computational cost. Possible future works include extending the embedding to userspace events, applying the models to downstream tasks such as anomaly detection, and applying the embedding to the many previous works that rely on LSTMs.

\section{ACKNOWLEDGMENT}

We would like to gratefully acknowledge the Natural Sciences and Engineering Research Council of Canada (NSERC), Prompt, Ericsson, Ciena, and EffciOS for funding this project.

% \bibliographystyle{IEEEtranN}
% \bibliography{bibliography.bib}
% Generated by IEEEtranN.bst, version: 1.14 (2015/08/26)

\end{document}